\documentclass{ecai} 

\usepackage{latexsym}
\usepackage{amssymb}
\usepackage{amsmath}
\usepackage{amsthm}
\usepackage{booktabs}
\usepackage{enumitem}
\usepackage{graphicx}
\usepackage{color}
\usepackage{cuted}
\usepackage{amsmath}
\usepackage{amsfonts}
\usepackage{listings}
\usepackage{xcolor}

\usepackage{algorithm}
\usepackage{algorithmic}
\usepackage{multirow}
\usepackage{graphicx}

\usepackage{graphicx}
\usepackage{graphicx}  

\usepackage{cuted}
\usepackage{amsmath}
\usepackage{amsfonts}
\usepackage{listings}
\usepackage{xcolor}

\usepackage{algorithm}
\usepackage{algorithmic}
\usepackage{multirow}
\usepackage{graphicx}

\usepackage{amsmath}
\usepackage{multirow}
\usepackage{graphicx}
\usepackage{stix,bbding,pifont,utfsym,fontawesome}

\definecolor{keyword}{rgb}{0,0,0}
\definecolor{comment}{rgb}{0,0,0}
\definecolor{string}{rgb}{0,0,0}
\usepackage{bbding}

\lstdefinestyle{mystyle}{
    backgroundcolor=\color{white},
    commentstyle=\color{comment},
    keywordstyle=\color{keyword},
    numberstyle=\tiny\color{gray},
    stringstyle=\color{string},
    basicstyle=\ttfamily\footnotesize,
    breaklines=true,
    captionpos=b,
    numbers=left,
    numbersep=5pt,
}

\usepackage{newfloat}
\usepackage{listings}

\newcommand{\BibTeX}{B\kern-.05em{\sc i\kern-.025em b}\kern-.08em\TeX}

\begin{document}


\begin{frontmatter}


\paperid{123} 


\title{{\em PoemTale Diffusion}: Minimising Information Loss in Poem to Image
Generation with Multi-Stage Prompt Refinement}



\author[A]{\fnms{Sofia}~\snm{Jamil}\thanks{Corresponding Author. Email: sofia\_2321cs16@iitp.ac.in}}
\author[A]{\fnms{Bollampalli Areen}~\snm{Reddy}}
\author[A]{\fnms{Raghvendra}~\snm{Kumar}} 
\author[A]{\fnms{Sriparna}~\snm{Saha}} 
\author[B]{\fnms{Koustava}~\snm{Goswami}}
\author[B]{\fnms{K. J.}~\snm{Joseph}}

\address[A]{Department of Computer Science \& Engineering, Indian Institute of Technology Patna, India}
\address[B]{Adobe Research}


\begin{abstract}
Recent advancements in text-to-image diffusion models have achieved remarkable success in generating realistic and diverse visual content. A critical factor in this process is the model's ability to accurately interpret textual prompts. However, these models often struggle with creative expressions, particularly those involving complex, abstract, or highly descriptive language. In this work, we introduce a novel training-free approach tailored to improve image generation for a unique form of creative language: poetic verse, which frequently features layered, abstract, and dual meanings.
Our proposed {\em PoemTale Diffusion} approach aims to minimise the information that is lost during poetic text-to-image conversion by integrating a multi stage prompt refinement loop into Language Models to enhance the interpretability of poetic texts. To support this, we adapt existing state-of-the-art diffusion models by modifying their self-attention mechanisms with a consistent self-attention technique to generate multiple consistent images, which are then collectively used to convey the poem's meaning. Moreover, to encourage research in the field of poetry, we introduce the {\em P4I (PoemForImage)} dataset, consisting of 1111 poems sourced from multiple online and offline resources. We engaged a panel of poetry experts for qualitative assessments. The results from both human and quantitative evaluations validate the efficacy of our method and contribute a novel perspective to poem-to-image generation with enhanced information capture in the generated images. 
\end{abstract}

\end{frontmatter}


\section{Introduction}


Poetry is a distinctive form of expression that evokes emotions and feeds the imagination. This variety makes reading poetry particularly captivating, as it invites readers to connect with the poet's inspiration while crafting their own personal perspectives. But what if these imagined scenes could be brought to life? Visualizing poetry adds a new dimension to the experience, allowing us to see and feel the poem while enhancing our imagination. With the rise of text-to-image generation models, particularly diffusion models, creative writing has seen remarkable advancements due to their ability to produce high-quality and diverse images. Several efforts have been made to tackle the zero-shot text-to-image generation challenge by pre-training large-scale generative models on massive image-text datasets, such as DALL-E \cite{dalle} and CogView \cite{cogview}. However, poetry, with its intricate layers of meaning and emotional depth, poses a unique challenge for accurate visual representation. Diffusion models often struggle with complex prompts that require a deeper level of comprehension. For example, even the latest SDXL \cite{sdxl_refiner} model frequently fails to generate an exact number of objects or correctly interpret negation in prompts \cite{lmd}.

This raises an important question: \textit{How can existing visualization models be adapted for poetry?}
One potential solution for poem-to-image generation is to curate a comprehensive multimodal dataset comprising poems and corresponding reference images and then train a text-to-image diffusion model to enhance poem understanding. However, this approach demands substantial time and resources to assemble a diverse and high-quality dataset, along with the huge computational cost of training or fine-tuning a diffusion model on such data. In contrast, we propose a novel, training-free solution that enables diffusion models to generate images that effectively capture the essence and meaning of a poem without the need for extensive retraining.

The quality of generated images in text-to-image diffusion models is inherently constrained by the text representation capabilities of their encoders. Most existing text-to-image models \cite{sdxl_refiner,dalle3,sdxl_turbo} rely on CLIP \cite{clip} to encode input prompts. However, CLIP's limited text representation capacity constrains the performance of text-to-image generative models. Taking this aspect into consideration, we propose an approach, {\textit PoemTale Diffusion}, that aims to address this limitation by focusing on three key areas for poetry-based image generation. First, since CLIP encoders struggle with complex poetic language, transforming poems into a more structured format improves image generation. To achieve this, we leverage LLM capabilities along with automatic prompt refinement techniques to convert poems into detailed image instructions, which are then used as prompts for image generation. Secondly, a single poem often contains significant variations in information, making it difficult to generate a single representative image. To capture the complex information within poems, we propose an entity-plus-emotion segmentation algorithm, which divides poems into distinct segments based on shifts in entities and emotional tones. Lastly, to generate visually coherent images while preserving a consistent flow of information throughout the poem’s depiction, we incorporate the consistent self-attention technique \cite{story_diff} into existing text-to-image generation models. To sum up, we make the following key contributions:
\begin{itemize}

\item We introduce a novel task for poem visualization, aiming to generate images that capture the most detailed information from poems.

\item We present the \textit{PoemTale Diffusion} Approach, which combines the poetic understanding of LLMs with automatic prompt refinement techniques and consistent self-attention to generate coherent and meaningful images.

\item We propose a novel poem segmentation algorithm, \textit{Entity Plus Emotion (EPE)}, which segments poems based on context shifts, capturing changes in entities and emotions to enhance poem clarity. 

\item We introduce a \textit{Poem for Image (P4I)} dataset, consisting of 1111 poems across various themes to assess the reliability of our approach and encourage future research in the field of poetry.

\item We conducted rigorous quantitative and human evaluations for each component of our pipeline, demonstrating that our proposed approach generates images with the maximum information retention.

\item All codes and datasets used in this study are publicly available at the GitHub \footnote{https://github.com/SofeeyaJ/PoemTale-Diffusion} repository.

\end{itemize}

\section{Background and Related Works}

\textbf{Image and Poetry: }
In the field of natural language processing (NLP), various generative tasks related to poetry have been extensively explored. These include topic-based poem generation \cite{topicpoem1, topicpoem2, topicpoem3}, acrostic poem generation \cite{acrostic}, and rhetorically controlled poetry generation \cite{rheotic}. The emergence of large language models (LLMs) has further led to significant research on different prompting techniques for poem generation \cite{llm1, llm2, llm3, jamil-etal-2025-poetry}. Some studies have also focused on generating poems from images using deep coupled visual-poetic embedding models combined with RNN-based adversarial training \cite{poemfromimage1}. In some works, poetry was generated from images using recurrent neural networks trained on existing poems \cite{poemfromimage2}. In contrast, \citet{li2021paint4poem} explored the generation of painting-like images for Chinese poems. Similarly, another study,  \citet{image_frompoem} focused on creating poetic Chinese landscape paintings with calligraphy. 

\textbf{Image Generation Via Diffusion Models:} The recent boom in diffusion models \cite{emn16} has fundamentally transformed the field of text-to-image generation. These models, such as IMAGEN \cite{emn13}, DALL-E 3 \cite{emn14}, and Stable Diffusion \cite{emn15}, have quickly become popular because of their exceptional capacity to generate images of high quality and variety. In a recent study by \citet{yu2024wonderjourney}, a text-driven point cloud generation pipeline was developed to create compelling and coherent sequences of 3D scenes. Similarly, \citet{liu2024intelligent} introduced a method to generate coherent image sequences based on a given storyline.  Moreover, some works have aimed to generate images while retaining specific identities. However, they require an initial reference image to maintain consistency across generated images. Several approaches have been developed for this purpose, such as DreamBooth \cite{emn21}, Custom Diffusion \cite{emn22}, IPAdapter \cite{emn23}, and PhotoMaker \cite{emn24}.  

\textit{However, existing literature lacks approaches for generating images that effectively convey a poem’s meaning, regardless of its genre, type, or style. Our work bridges this gap by introducing the PoemTale Diffusion approach, which produces a sequence of descriptive images.}


\section{P4I Corpus}
\label{annotationsd}


To assess the scalability, robustness, and reliability of our framework, we required a dataset that covers a wide range of thematic content. To achieve this, we curated a diverse selection of poem types that are not commonly found in existing literature. We refer to this dataset as \textit{P4I (\textbf{P}oems\textbf{ForI}mages)}. We employed three undergraduate Computer Science and Engineering students and one postgraduate humanities student from a reputable university. These individuals were carefully selected based on their technical proficiency in collecting and organizing the required poems from various sources. The students were compensated for their efforts \footnote{The annotators received compensation in the
form of gift vouchers and honorarium amounts in accordance with

\url{https://www.minimum-wage.org/}}, and the entire process was completed in 270 hours.

\subsection{Data Collection}

\begin{enumerate}

 \item Online Sources: To construct the \textit{PoemforImage(P4I)} dataset, we gathered poems from multiple sources, including high-quality online platforms such as Poetry Foundation \footnote{\url{https://poemanalysis.com/}} and Poem Analysis\footnote{\url{https://www.poetryfoundation.org/}}. The poems span a variety of themes, such as narrative, free verse, odes, laments, open verse, sonnets, open couplets, and children's poetry. In response to the digital age's influence on poetry, we also included modern forms like internet-based poetry and slam poetry.
 \item Offline Sources: We manually curated poems from books, journals, and articles available in offline formats, sourcing them from three renowned libraries in the state.

\end{enumerate}


\subsection{Data Annotation}
We annotated the P4I dataset with structured metadata to improve poem organization and facilitate image generation. We enlisted three PhD students from the Department of English Literature to conduct the annotation, following guidelines curated by a professor in Literary Arts. All annotators were duly compensated for their contributions and will be acknowledged in the research. To ensure consistency and reliability, inter-annotator agreement was measured using Cohen’s Kappa score \cite{kappa}, achieving a value of 0.78, which confirms the consistency of the annotation process. Table \ref{p4i} presents the various statistics of our P4I dataset.

\subsection{Annotation Guidelines}

\begin{itemize}

 \item  Remove duplicate records, extraneous characters, unnecessary white spaces, corrupt values, and HTML tags. 
 \item  Standardize text formatting and resolve any inconsistencies in poem structure.
 \item  Label the Poem Title, Poet, and Poem Theme for each entry in the dataset.
 \item  Identify and label the protagonist or central character of the poem.
 \item  Identify the predominant mood or emotion of the poem. If the poem exhibits an emotional shift, segment the poem into smaller verses or stanzas corresponding to these shifts. Label each segment with its predominant emotion.

\end{itemize}






\begin{table}[!htpb]
\caption{Statistics of Different Poems Present in P4I Dataset}
\centering
\resizebox{0.85\columnwidth}{!}{%
\begin{tabular}{lc}
\hline
\multicolumn{1}{c}{Statistics} & Count \\ \hline
\textit{Total Number of Poems} & 1111 \\
\textit{Maximum Length of Poems (In words)} & 460 \\
\textit{Minimum Length of Poems (In words)} & 16 \\
\textit{Average Poem Length (in words)} & 180 \\
\textit{Total Themes of poems} & 6 \\
\textit{Distinct number of Poets} & 798 \\ \hline
\end{tabular}%
}
\label{p4i}
\end{table}
\section{Tasks Setups}

Poetry evokes vivid thoughts in the minds of its readers. The process of visualizing and interpreting it is inherently subjective, as each individual perceives it differently. No single representation can fully capture every detail, which presents an interesting challenge when converting these meanings into images. To address this challenge, we explored different approaches for generating images from poems and proposed a solution that preserves their essence while minimizing information loss during the transformation process.

\subsection{Approach 1: Image Generation via Direct Poems}

We formulate this task as a simple text-to-image generation problem, where we utilize SDXL \cite{sdxl_refiner} to transform poetry into engaging visuals that capture the poem's meaning and theme. We take the entire poem as input and provide it to diffusion models to generate an image corresponding to the input poem. 




\subsection{Approach 2: Image Generation per Poem Stanzas }


As per the annotation process highlighted in Section \ref{annotationsd}, poems were segmented whenever there was a change in emotion. Therefore, we took those segments and fed them into the diffusion model to generate a corresponding image. This approach was designed to maximize information capture and ensure that the visuals remained closely aligned with the poem.



\subsection{Analysis of Approaches}

For the analysis of the above approaches, we collaborated with a team of four professional experts from the Poetry Society of the Nation, comprising one renowned poet, two university professors, and a doctoral researcher in visual and literary arts. We randomly selected 15\% of the samples from the P4I dataset, totaling 170 poems. To ensure a comprehensive evaluation, the assessment was conducted based on two key factors: \textit{(i) semantic alignment}, which measures how well the images correspond to the poem’s meaning, \textit{(ii) emotional resonance}, which examines the extent to which the visuals capture the intended emotions of the poem. Each expert provided individual ratings for these factors per poem, and the final scores were combined to determine overall performance. To ensure an unbiased evaluation, the judges were not informed of the source of the generated images during the rating process. Table \ref{humanevaluation} presents the average expert scores for this task. The results indicate that Approach 1 caused significant information loss, as a single image failed to capture the poem’s meaning, highlighting the limitations of current image generation models in interpreting and visualizing poetic language. Similarly, in Approach 2, where multiple images were generated for different stanzas, the generated images lacked semantic alignment, leading to fragmented information capture.

\section{Room for Improvement: {\em PoemTale} Diffusion Approach}


In light of the experimental results and analysis, it is evident that both approaches failed to generate descriptive and meaningful visuals when processed by text-to-image models. These models produce images that lack depth, coherence, and alignment with the poem’s meaning, resulting in significant information loss. To address these limitations in a resource-efficient manner, we propose the {\em PoemTale Diffusion} Approach to enhance information capture, focusing on the key areas below for improvement.




\textbf{1.} Restructuring poems into a format more understandable to text encoders by clarifying abstract language, expanding implicit details, and simplifying complex metaphors can improve the model’s ability to accurately interpret poetic meaning. 

\textbf{2.} To maintain narrative flow, the main character or central theme of the poem should remain visually consistent throughout all generated images.

\textbf{3.} Some poems contain long, interconnected verses that shift in tone, theme, or setting. Instead of processing an entire poem as a single prompt, we propose segmenting it into smaller, distinct parts.

\begin{figure}[!ht]
\centerline{\includegraphics[width=\columnwidth]{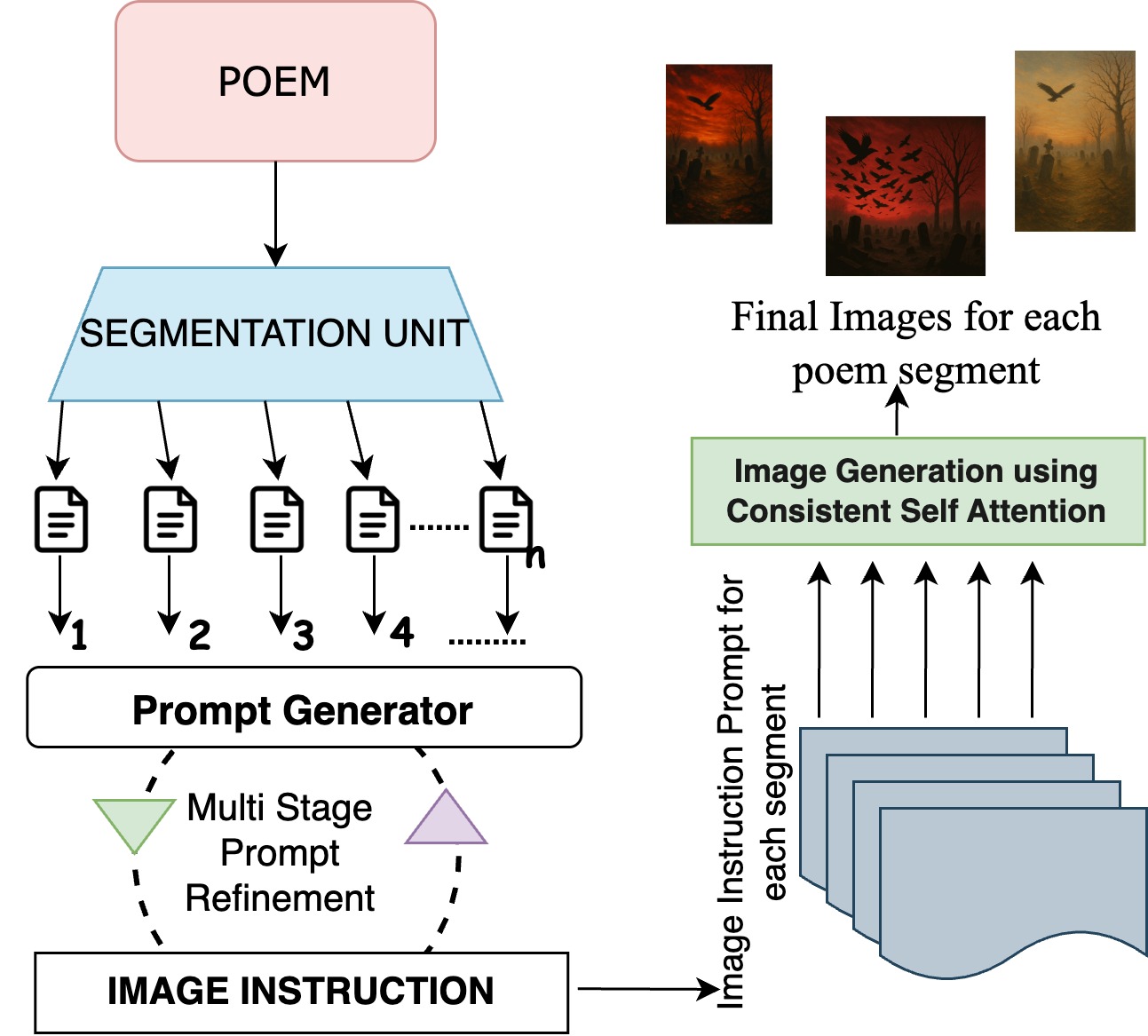}}
\caption{Workflow of our proposed \textit{PoemTale Diffusion} approach}
\label{workflow}
\end{figure}

To effectively implement the above strategies, we introduce \textit{PoemTale Diffusion}, a novel approach that generates a sequence of images, \( \{I_{S_1}, I_{S_2}, \dots, I_{S_n}\} \), where each image \( I_{S_i} \) visually represents the meaning of its respective segment \( S_i \). The collective sequence \( \{I_{S_1}, I_{S_2}, \dots, I_{S_n}\} \) of images captures the narrative structure and collective meaning of the entire poem, preserving both semantic depth and emotional resonance. This method enhances poetry-to-image alignment while minimizing poetic information loss. Our approach is divided into three distinct modules: (i) Segmentation Module, (ii) Prompt Generator incorporating Multi Stage Prompt Refinement (MSPR) loop (iii) Text to Image Models. Figure \ref{workflow} depicts the working flow of our proposed approach. 

\subsection{Segmentation Module}
To effectively break down complex poetic verses into smaller, independent units, we introduce an \emph{Entity Plus Emotion (EPE)} approach to detect segment boundaries in a poem. Our approach incrementally processes each line of the poem, assigning named entity recognition (NER) tags and emotions to each line. When there is a change in NER tags or emotions between consecutive lines, the algorithm detects a shift in context and segments the poem accordingly. This method focuses exclusively on identifying where one segment of the poem ends, and another begins, based on changes in the emotion and entities. We employed the NLTK library to identify entities in the poem and employed a fine-tuned version of DistilRoBERTa-base \cite{emotion_citation} for the classification of emotions.

\subsection{Multi Stage Prompt Refinement Loop (MSPR)}

\begin{figure}[!ht]
\centerline{\includegraphics[width=\columnwidth]{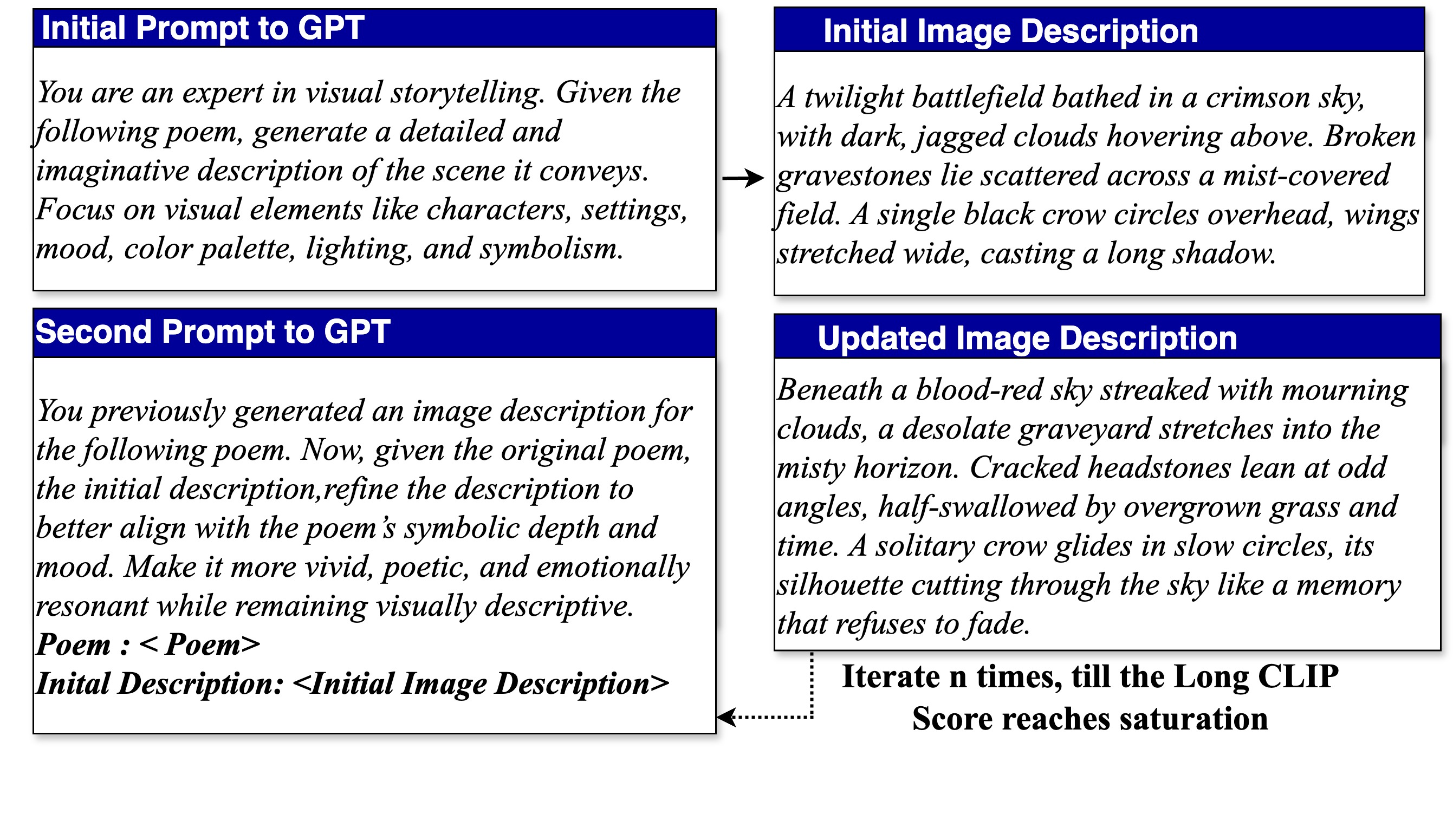}}
\caption{Working of Multi Stage Prompt Refinement Approach}
\label{msp}
\end{figure}

To clarify abstract language, reveal hidden details, and simplify complex metaphors in poems, we utilized a multi-stage prompt refinement loop leveraging \texttt{GPT-4o-mini} for generating image descriptions. In the multi-stage refinement loop, the process begins with Stage 1 \cite{jamil-etal-2025-poetry}, where GPT is prompted as a visual storytelling expert to generate a vivid and imaginative scene based solely on a given poem. The description is evaluated using the Long CLIP \cite{longclip} score. In Stage 2, refinement is introduced: As demonstrated in Figure \ref{msp}, GPT receives the original poem along with the initial description and is prompted to deepen the emotional and poetic resonance of the scene. This stage encourages symbolic interpretation and more detailed visualization to align the output closely with the poem's main intent. This iterative process continues, with each cycle aiming to improve description alignment with the poem's meaning. The loop terminates when the Long CLIP score shows saturation, indicating that there is no significant improvement across three consecutive iterations, showing convergence. To supplement automated evaluation, we also integrated feedback from human experts. However, due to the time and resource demands of manual assessment, this method was applied to a smaller subset of poems. Across these samples, we consistently observed convergence after 4–5 iterations, which we identify as the optimal point for generating image descriptions from poetic input. It is to be noted that this multi-stage refinement is applied to all the segments of the poems to generate accurate image descriptions for each segment. These image descriptions for each segment are utilized in the next step to form a visual story for each poem. This MSPR requires absolutely no training time and is ideal for real-time scenarios.

\begin{table*}[!htpb]
\caption{The Table presents the performance of different text-to-image generation models evaluated under various approaches. The best-performing approach values are highlighted in bold, while within each of the other approaches, the best-performing model is italicized. `/' denotes results are not applicable for that metric.}
\centering
\resizebox{\textwidth}{!}{%
\begin{tabular}{llcccc}
\hline
 &  & \multicolumn{2}{c}{\textbf{PROMPT GENERATOR UNIT}} & \textbf{EMOTION} & \textbf{CHARACTER CONSISTENCY} \\
 &  & BLIP & Long-CLIP & CLIP & CLIP \\ \hline
\multirow{3}{*}{\textbf{\begin{tabular}[c]{@{}l@{}}PoemTale Diffusion\\ Approach\end{tabular}}} 
 & \textit{JANUS} & \textbf{0.4009} & \textbf{0.3928} & \textbf{0.4028} & \textbf{0.2184} \\
 & \textit{SDXL} & \textbf{0.4218} & \textbf{0.4605} & \textbf{0.3926} & \textbf{0.2859} \\
 & \textit{PLAYGROUND V3} & \textbf{0.4333} & \textbf{0.5897} & \textbf{0.4249} & \textbf{0.3070} \\ \hline
\multirow{3}{*}{\textbf{\begin{tabular}[c]{@{}l@{}}Only With Poem Segments\\ (Approach 2)\end{tabular}}} 
 & \textit{JANUS} & 0.2066 & 0.1808 & 0.2355 & 0.1193 \\
 & \textit{SDXL} & 0.3306 & 0.2464 & 0.2328 & 0.1864 \\
 & \textit{PLAYGROUND V3} & \textit{0.3969} & \textit{0.2567} & \textit{0.2383} & \textit{0.1948} \\ \hline
\multirow{3}{*}{\textbf{\begin{tabular}[c]{@{}l@{}}Single Image for Poems\\ (Approach 1)\end{tabular}}} 
 & \textit{JANUS} & 0.1845 & 0.1688 & 0.2096 & / \\
 & \textit{SDXL} & 0.2846 & 0.2131 & 0.1692 & / \\
 & \textit{PLAYGROUND V3} & \textit{0.3224} & \textit{0.2193} & \textit{0.2145} & / \\ \hline
\end{tabular}%
}
\label{quantitative_evaluation}
\end{table*}
\subsection{Poem to Image Generation using Consistent Self-Attention}
After we have detailed image instructions for all segments of the poem, our final step involves generating images for multiple segments (if present) of a poem using a diffusion model. To maintain coherence across multiple images, the main character or central theme of the poem should remain visually consistent throughout all generated images. As demonstrated in previous research \cite{attention}, self-attention plays a crucial role in modeling the overall structure of generated visual images. Therefore, we incorporate Consistent Self-Attention \cite{story_diff}, which can be seamlessly integrated into any diffusion model by replacing the original self-attention layer in a zero-shot manner. This approach is training-free and model-independent, making it applicable to any text-to-image generation model equipped with a self-attention layer. The process begins with the generation of an initial image for the first segment of the poem. Subsequently, sampled reference tokens from the initial image are utilized during the token similarity matrix calculation and token merging. Since these sampled tokens share the same $Q-K-V$ (Query-Key-Value) weights, no additional training is required. As shown in Figure \ref{ablation}, the application of Consistent Self-Attention successfully preserves identity and attire consistency, ensuring coherence across all generated images. 

Consistent Self-Attention is integrated into the U-Net architecture of existing diffusion models, utilizing the original self-attention weights to remain training-free. Given a batch of image features $I \in \mathbb{R}^{B \times N \times C}$, where $B$ is the batch size, $N$ is the number of tokens in each image, and $C$ is the channel count, the self-attention function is defined as:

\begin{equation}
O_i = \text{Attention}(Q_i, K_i, V_i)
\end{equation}

where $Q_i, K_i, V_i$ represent the query, key, and value matrices, respectively. Standard self-attention operates independently within each image feature $I_i$.

To enforce subject consistency across images, Consistent Self-Attention samples tokens from other images within the batch:

\begin{equation}
S_i = \text{RandSample}(I_1, I_2, I_{(i-1)}, I_{(i+1)}, ...)
\end{equation}

where $\text{RandSample}$ denotes the random sampling function. These sampled tokens $S_i$ are then paired with the original image feature $I_i$ to form a new token set $P_i$. Linear projections generate the updated key $K_{P_i}$ and value $V_{P_i}$ matrices for Consistent Self-Attention, while the original query $Q_i$ remains unchanged. The final attention operation is computed as:

\begin{equation}
O_i = \text{Attention}(Q_i, K_{P_i}, V_{P_i})
\end{equation}

By facilitating cross-image interactions, Consistent Self-Attention ensures that character identity, facial features, and attire remain consistent across multiple images. This technique ultimately improves poetic information preservation by aligning generated images more closely with the poem’s intended meaning.

\section{Experiments}

\subsection{Baselines}

We implemented our \textit{PoemTale Diffusion} approach using three State-of-the-art (SOTA) text-to-image generation models: Playground V3 \cite{playgroundv3}, Stable Diffusion XL \cite{sdxl_refiner}, and Janus \cite{janus}. These models were chosen for their training-free and pluggable nature, allowing us to establish baselines and compare results across different architectures.
For efficient sampling, we employed DDIM (Denoising Diffusion Implicit Models) to accelerate the inference process, INT8 quantization to optimize memory usage, and EMA (Exponential Moving Average) for checkpoint selection.

\subsection{Evaluation Metric}

Since our \textit{PoemTale Diffusion} approach comprises three distinct modules, it is essential to evaluate them individually to assess the errors introduced at each stage and their overall impact on image quality. To achieve this, we utilized the CLIP score and BLIP score to evaluate each component of {\em PoemTale Diffusion}. To evaluate the alignment between the generated image and its corresponding prompt, we utilized the BLIP \cite{blip} Score, where captions were generated for images, and their similarity to the original image instruction prompt was computed. Additionally, Long-CLIP \cite{longclip} was employed to measure the cosine similarity between the poem and the generated image. Emotion plays a crucial role in conveying a poem’s intended meaning. To assess how well the generated image captures the poem’s emotional nature, we compared the emotion depicted in the image with the gold reference emotion expressed in the poem, using CLIP scores \cite{clip} for measurement. This detailed assessment of each component allows us to analyze the extent of information loss at each step and its overall impact on the final generated image, ensuring a thorough evaluation of our approach.

\begin{figure}[!htpb]
\centerline{\includegraphics[width=\columnwidth]{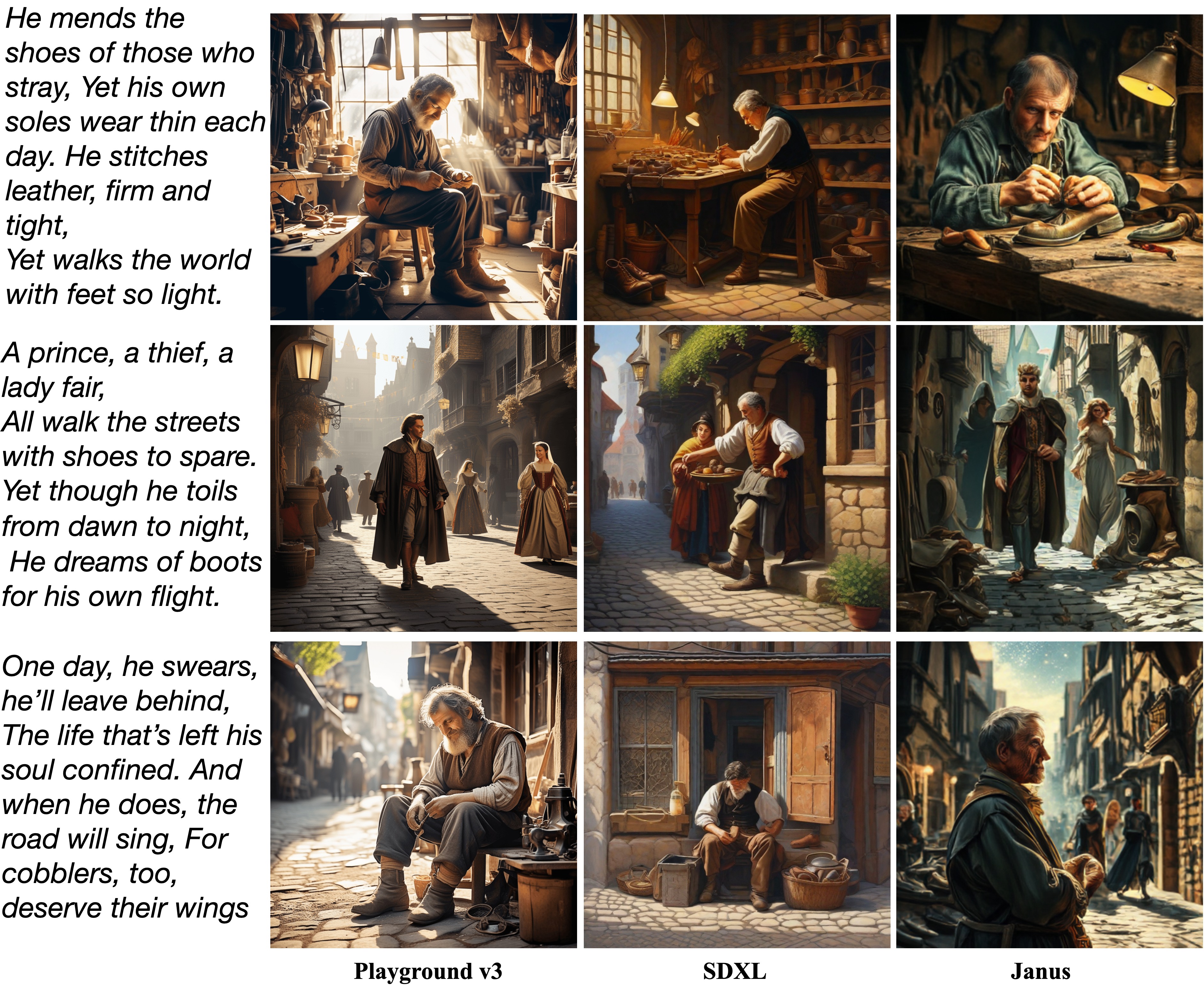}}
\caption{Qualitative Comparison of Different Image Generation Models Integrated with \textit{PoemTale Diffusion} approach. We recommend that the reader zoom in for better readability.}
\label{allmodels}
\end{figure}

\begin{figure*}[!htbp]
\includegraphics[width=0.9\textwidth]{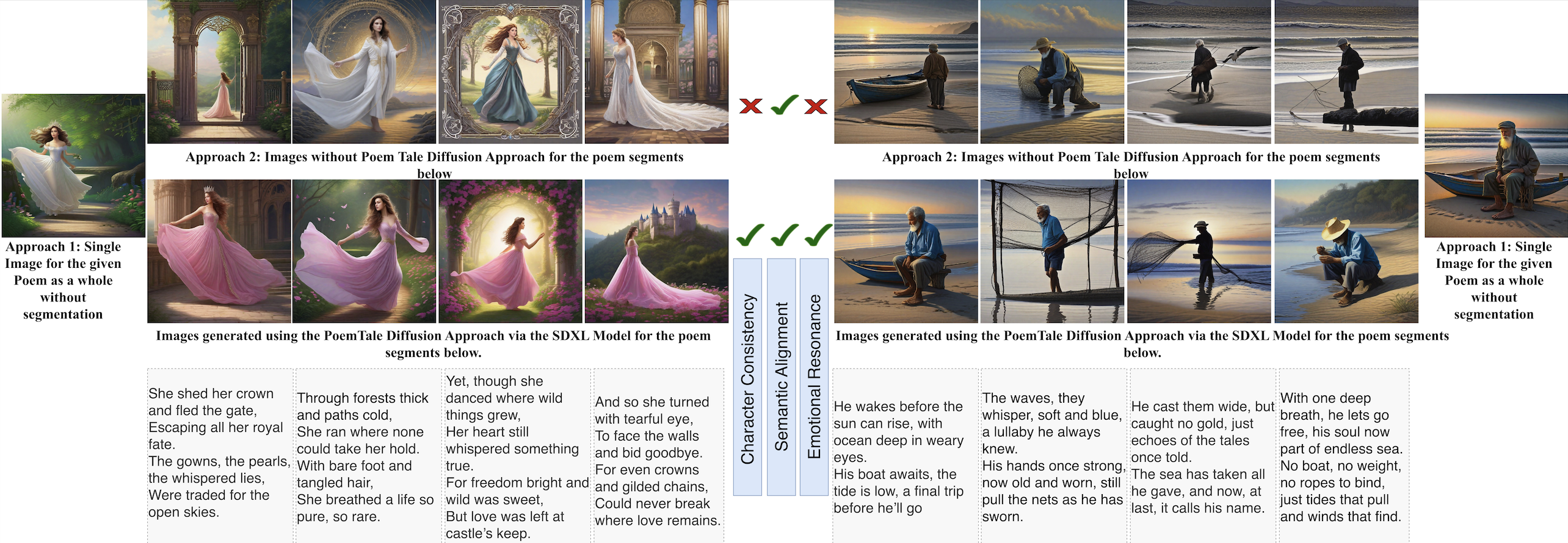}
\caption{Results of various approaches in poem to image generation corresponding to various segments of the poem (shown in the grey boxes). We recommend that the reader zoom in for better readability.}
\label{ablation}
\end{figure*}

\section{Results Analysis and Discussion}

\subsection{Quantitative Evaluation}

The results detailed in Table \ref{quantitative_evaluation} provide important insights, into different approaches within the context of our task. We conducted a quantitative evaluation based on two key factors: 
\textit{(i) Approach-Based Evaluation:} As shown in Table \ref{quantitative_evaluation}, the \textit{PoemTale Diffusion} Approach consistently outperforms all other methods for poem-to-image generation. This highlights the importance of integrating poem segmentation and consistent self-attention, which together result in high-quality, thematically aligned, and more informative images. We observe an increasing trend in CLIP scores across the following aspects: emotion depiction (0.21$\uparrow$), prompt alignment with images (1.68$\uparrow$), character consistency across images (0.57$\uparrow$), and resemblance to the reference character in the poem (0.82$\uparrow$).

\textit{(ii) Baseline Based Evaluation:} We implemented our proposed approach across different baseline models and conducted both qualitative and quantitative evaluations. Figure \ref{allmodels} presents a qualitative comparison of the generated images. Playground V3, when integrated with the \textit{PoemTale Diffusion} approach, produces the most detailed and most visually appealing images.  It successfully maintains both facial features and attire, ensuring character consistency.  Other models show improved character coherence but occasionally fail to preserve finer details like clothing and hairstyle.  Yet across all models, the use of \textit{PoemTale Diffusion} produces significantly more thematically aligned and visually coherent images compared to approaches that do not utilize it.

\begin{figure}[!ht]
\centerline{\includegraphics[width=\columnwidth]{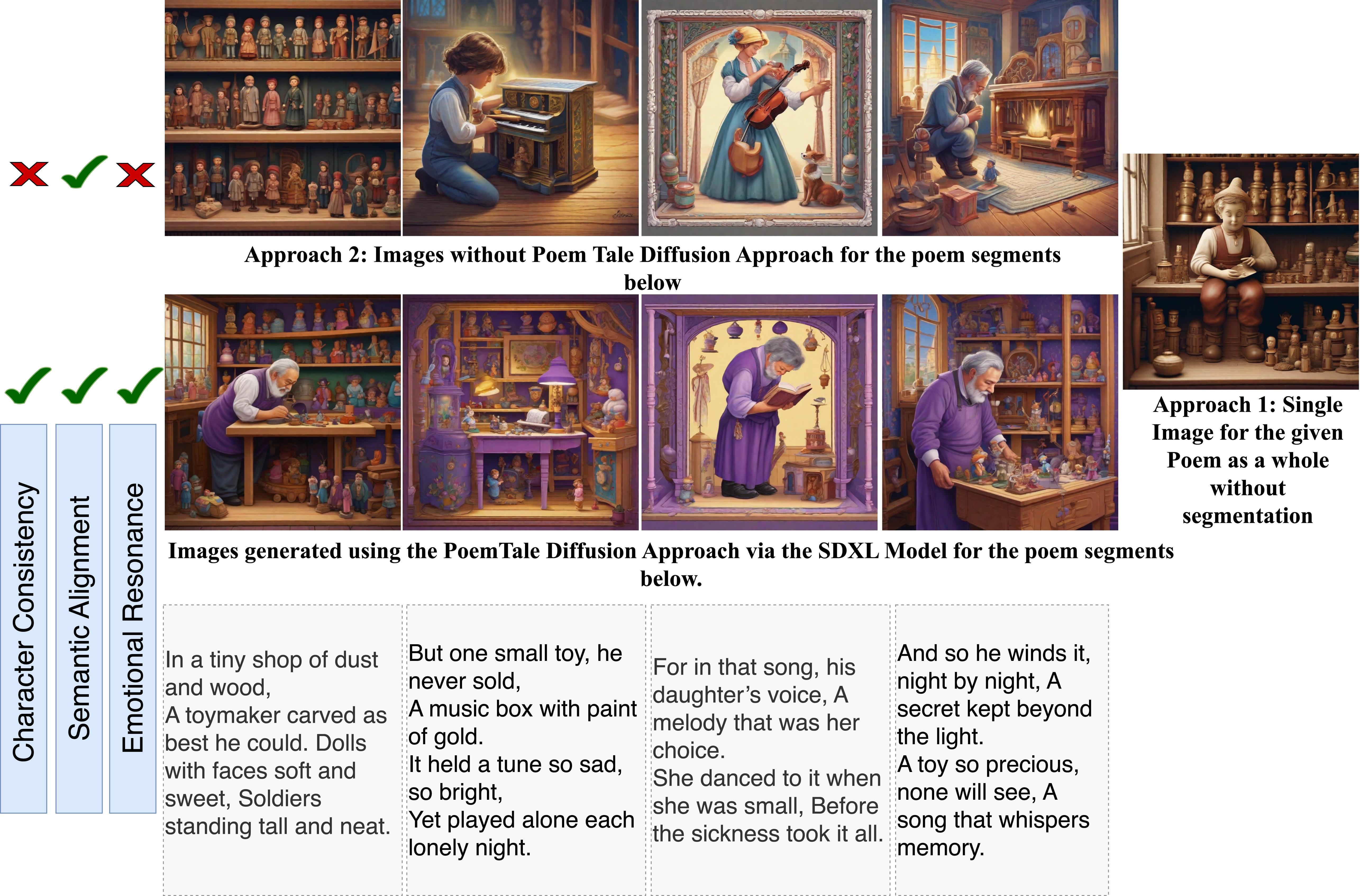}}
\caption{Results of various approaches in poem to image generation corresponding to various segments of the poem (shown in the grey boxes). We recommend that the reader zoom in for better readability.}
\label{appendiz01}
\end{figure}

\begin{figure}[!ht]
\centerline{\includegraphics[width=\columnwidth]{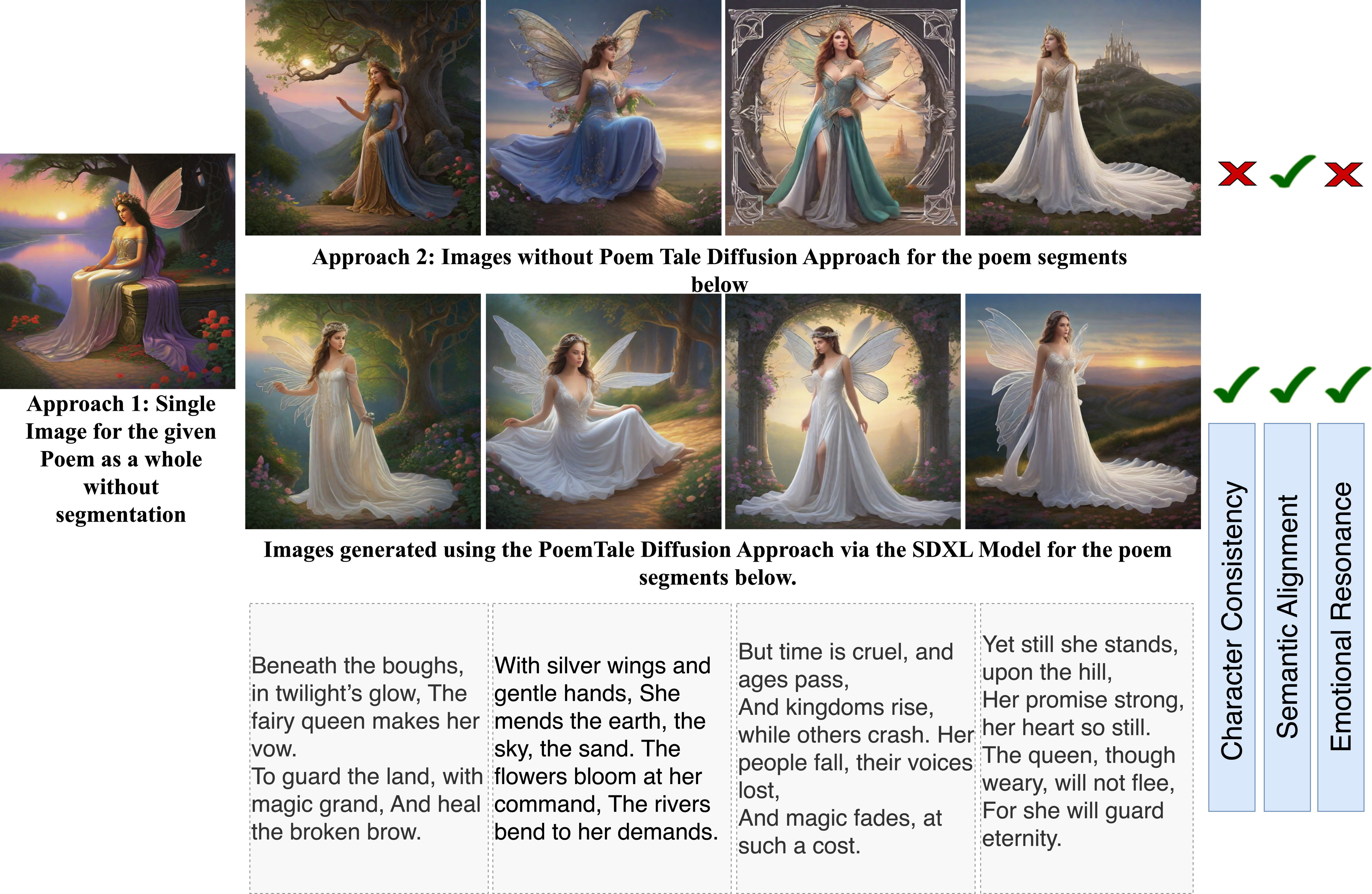}}
\caption{Results of various approaches in poem to image generation corresponding to various segments of the poem (shown in the grey boxes). We recommend that the reader zoom in for better readability.}
\label{appendix02}
\end{figure}


\begin{table}[!htpb]
\caption{Human evaluation scores from various experts for different approaches, on a randomly selected set of 150 poems.}
\centering
\resizebox{\columnwidth}{!}{%
\begin{tabular}{llc}
\hline
 &  & Average Expert Scores \\ \hline
\multirow{2}{*}{\begin{tabular}[c]{@{}l@{}}{\em PoemTale Diffusion} \\ Approach\end{tabular}} & \textit{Semantic Alignment} & \textbf{3.9} \\
 & \textit{Emotional Resonance} & \textbf{4.1} \\ \hline
 \multirow{2}{*}{\begin{tabular}[c]{@{}l@{}}Approach 1:\\ \textit{Single Image for Poems}\end{tabular}} & \textit{Semantic Alignment} & 1.8 \\
 & \textit{Emotional Resonance} & 1.7 \\ \hline
\multirow{2}{*}{\begin{tabular}[c]{@{}l@{}}Approach 2:\\ \textit{Only With Poem Segments}\end{tabular}} & \textit{Semantic Alignment} & 2.2 \\
 & \textit{Emotional Resonance} & 2.1 \\ \hline
\end{tabular}%
}
\label{humanevaluation}
\end{table}



\subsection{Human Evaluation}

Recognizing that existing metrics may not fully capture the quality of the generated images, especially in terms of consistency, relevance, and aesthetic appeal, and with no standardized metric available, we incorporated human evaluations. To conduct this evaluation, we collaborated with the same set of area experts from the Poetry Society of the Nation, including two renowned poet, one university professors, and a doctoral researcher in visual and literary arts. We selected 15\% of the samples from the \textit{P4I} dataset. The experts reviewed the images produced by our \textit{PoemTale Diffusion} approach using the playground model (as it is the best performing baseline). For each selected poem, experts were asked to rate the images on a scale from 1 to 5, considering the criteria of image-poem semantic alignment, Consistency, and Emotional Resonance, with higher scores reflecting better quality. The final rating for each image across all categories was determined by averaging the scores provided by the three experts. To eliminate potential bias, the experts were not informed of the names of the models used to generate the images or the approaches taken for image generation. The human evaluation results, as shown in Table \ref{humanevaluation}, indicate that our \textit{PoemTale Diffusion} approach achieves superior performance in overall scores, particularly in terms of consistency and information retention.

Additionally, we present a detailed analysis of the images produced by different approaches in Figure \ref{ablation}, based on the same key aspects used for human evaluation. \textit{(I) character consistency:} The \textit{PoemTale Diffusion} approach has an impressive ability to maintain character consistency across all image segments. Figure \ref{appendix02} shows that the generated images consistently depict the main character with the same facial features and attire throughout, whereas other approaches fail to preserve this consistency across subsequent images. This improvement can be attributed to the Consistent Self-Attention mechanism, which effectively recognize and maintain the character's identity across all generated images.
\textit{(ii) semantic alignment:} Figure \ref{ablation} illustrates that our \textit{PoemTale Diffusion} approach achieves the highest level of poem-to-image alignment. This improvement is due to the Prompt Generator Unit, which restructures the poem into a more understandable format by generating detailed image instructions. Furthermore, the MSPR technique ensures that the final image instructions accurately convey the poem's meaning, resulting in highly aligned visual representations. \textit{(iii) emotional resonance:} Since poems often convey shifting emotions, we segmented them based on emotional transitions and key entities. As seen in Figure \ref{appendiz01}, a single image struggles to capture the full emotional depth of a poem. However, by generating images for individual segments based on emotional shifts, our approach ensures that each image reflects the intended mood of its corresponding poem segment. 

\section{Conclusion and Future Works}

This work explores the capabilities of diffusion models in poem-to-image generation by evaluating various traditional approaches based on semantic relevance and emotional resonance. To address the limitations of existing methods, we introduce the \textit{PoemTale Diffusion} approach, which requires no training and is designed to generate images that capture the maximum amount of information and meaning from poems. We have implemented our proposed method on several SOTA diffusion models and thoroughly evaluated each component of the pipeline to assess how results propagate through to the final stage. To assess the scalability and effectiveness of our approach, we present the \textit{PoemForImage (P4I)} dataset, which consists of 1111 poems covering a diverse range of themes and styles. Our approach is the first to focus on image generation for diverse poetry, producing both single and multiple images. Looking ahead, we aim to develop a language-agnostic visual model capable of generating images for poetry across multiple languages.

\section{Acknowledgement}

Sriparna Saha would like to acknowledge the funding from ADOBE Research for conducting this research.

\bibliography{mybibfile}

\end{document}